\documentclass{article}

\usepackage[final]{corl_2021} % Uncomment for the camera-ready ``final'' version
\usepackage{microtype}
\usepackage{graphicx}
\usepackage{mathtools}
\usepackage{siunitx}
\usepackage{dsfont}
\usepackage{multirow}
\newcolumntype{?}[1]{!{\vrule width #1}}
\usepackage[font=small]{caption}
\usepackage{booktabs} % for professional tables
\usepackage{amssymb}
\usepackage{amsmath,amsthm}
\usepackage{mathtools}
\usepackage{subcaption}
\usepackage{algorithm}
\usepackage[noend]{algpseudocode}
\usepackage[utf8]{inputenc}
\usepackage[english]{babel}
\usepackage{caption}
\usepackage{wrapfig}
\usepackage{subcaption}
\usepackage{enumitem}
\usepackage{xcolor}

\theoremstyle{definition}

\usepackage{hyperref}
\usepackage{cleveref}
\newcommand{\todo}[1]{}
\renewcommand{\todo}[1]{{\color{red} TODO: {#1}}}

\DeclareMathOperator*{\argmin}{arg\,min}

\newcommand{\incmtt}[1]{{\fontfamily{cmtt}\selectfont{#1}}} % typewriter style
\newcommand{\new}[1]{{#1}}
\definecolor{darkblue}{rgb}{0.0,0.0,0.7} % for hyper-links
\renewcommand\algorithmiccomment[1]{%
  \textcolor{darkblue}{\scriptsize \incmtt{\hfill\#\,{#1}}}%
}

\usepackage{titlesec}
\titlespacing\section{0pt}{0pt plus 2pt minus 2pt}{0pt plus 2pt minus 2pt}
\titlespacing\subsection{0pt}{3pt plus 4pt minus 2pt}{0pt plus 2pt minus 2pt}
\titlespacing\subsubsection{0pt}{3pt plus 4pt minus 2pt}{0pt plus 2pt minus 2pt}

\usepackage{xspace}
\usepackage{amssymb}
\usepackage{amsmath}

\title{\LARGE \bf
\new{ThriftyDAgger:} Budget-Aware Novelty and Risk Gating for Interactive Imitation Learning
}

\def \algnamefull{ThriftyDAgger\xspace}
\def \algname{ThriftyDAgger\xspace}

\newcommand{\cloningloss} {\mathcal{L}(\pi_{r}(s_t),\pi_{h}(s_t))}

\newcommand{\tausup} {\beta_{h}} % tau? textrm?
\newcommand{\tauauto} {\beta_{r}} % textrm auto?
\newcommand{\deltasup} {\delta_{h}}
\newcommand{\deltaauto} {\delta_{r}}
\newcommand{\pisup} {\pi_{h}} 
\newcommand{\pirob} {\pi_{r}}
\newcommand{\safety} {\hat Q_{\phi, \mathcal{G}}^{\pi_r}}
\newcommand{\pimeta} {\pi}
\newcommand{\alphasupdesired} {\alpha_{h}}
\newcommand{\indicatorint} {m_I} % Daniel S: is this what he wanted?
\author{
  Ryan Hoque$^1$, Ashwin Balakrishna$^1$, Ellen Novoseller$^1$, \\
  \textbf{Albert Wilcox}$^1$, \textbf{Daniel S. Brown}$^1$, \textbf{Ken Goldberg}$^1$
  \thanks{$^{1}$AUTOLAB at the University of California, Berkeley}
  \thanks{Correspondence to {\tt\small ryanhoque@berkeley.edu}}
}
\makeatletter
\def\thanks#1{\protected@xdef\@thanks{\@thanks
        \protect\footnotetext{#1}}}
\makeatother
\begin{document}
\maketitle

%===============================================================================

\begin{abstract}
%Interventions are an intuitive way for a human to convey information about desired behaviors to a robotic agent and improve the performance of the deployed robot policy. However, such interventions cost
\new{Effective robot learning often requires online human feedback and interventions that can cost} significant human time, giving rise to the central challenge in interactive imitation learning: \textit{is it possible to control the timing and length of interventions to both facilitate learning and limit burden on the human supervisor?}
\new{This paper presents} \algnamefull, an algorithm for actively querying a human supervisor given a desired \new{budget of human interventions}. \algname uses a learned switching policy to solicit interventions only at states that are sufficiently (1) \textit{novel}, where the robot policy has no reference behavior to imitate, or (2) \textit{risky}, where the robot has low confidence in task completion. \new{To detect the latter, we introduce a novel metric for estimating risk under the current robot policy.} Experiments in simulation and on a physical cable routing experiment suggest that \algname's \new{intervention criteria} balances task performance and supervisor burden more effectively than prior algorithms. \new{ThriftyDAgger can also be applied at execution time, where it achieves} a $100\%$ success rate on both the simulation and physical tasks. A user study ($N=10$) in which users control a three-robot fleet while also performing a concentration task suggests that \algname increases human and robot performance by $58\%$ and $80\%$ respectively compared to the next best algorithm while reducing supervisor burden. See \url{https://tinyurl.com/thrifty-dagger} for supplementary material.

\end{abstract}

% Two or three meaningful keywords should be added here
%===============================================================================

\section{Introduction}
\label{sec:introduction}
% Motivate the problem's importance at a high level, discuss the central challenge in interactive IL (when to solicit interventions) and a brief high-level idea of the spectrum of different approaches (low supervisor burden, low performance, high supervisor burden, high performance)
Imitation learning (IL)~\cite{argall2009survey,invitation-imitation,osa2018algorithmic} has seen success in a variety of robotic tasks ranging from autonomous driving~\cite{pomerleau1991efficient,agile_driving,codevilla2018end} to robotic manipulation~\cite{one-shot-visual-IL,fang2019survey,ganapathi2020learning,rope-descriptors,kroemer2021review}. In its simplest form, the human provides an offline set of task demonstrations to the robot, which the robot uses to match human behavior. However, this offline approach can lead to low task performance due to a mismatch between the state distribution encountered in the demonstrations and that visited by the robot~\cite{dagger, DART}, resulting in brittle policies that cannot be effectively deployed in real-world applications~\cite{rlblogpost}. \textit{Interactive imitation learning}, in which the robot periodically cedes control to a human supervisor for corrective interventions, has emerged as a promising technique to address these challenges~\cite{hg_dagger,EIL, jauhri2020interactive,lazydagger}. However, while interventions make it possible to learn robust policies, these interventions require significant human time. Thus, the central challenge in interactive IL algorithms is to control the timing and length of interventions to balance task performance and the burden imposed on the human supervisor~\cite{safe_dagger,lazydagger}. Achieving this balance is even more critical if the human supervisor must oversee multiple robots at once~\cite{crandall2005validating,chen2010supervisory,swamy2020scaled}, for instance supervising a \new{fleet of robots in a warehouse~\cite{qt-opt} or self-driving taxis~\cite{codevilla2018end}. Since even relatively reliable robot policies inevitably encounter new situations that must fall back on human expertise, this problem is immediately relevant to contemporary companies such as Waymo and Plus One Robotics.}
% Companies such as Zoox, Nimble Robotics, Kiwibot, and Plus One Robotics seek to automate everything from self-driving to bin-picking with robot fleets that fall back on the expertise of human teleoperators when necessary.}

% Why is this challenging in the context of prior work:
One way to determine when to solicit interventions is to allow the human supervisor to decide when to provide the corrective interventions. However, these approaches---termed ``human-gated" interactive IL algorithms~\cite{hg_dagger, EIL, HITL}---require the human supervisor to continuously monitor the robot to determine when to intervene. This imposes significant burden on the supervisor and cannot effectively scale to settings in which a small number humans supervise a large number of robots. To address this challenge, there has been recent interest in approaches that enable the robot to actively query humans for interventions, called ``robot-gated" algorithms~\cite{safe_dagger,SHIV, ensemble_dagger, lazydagger}. Robot-gated methods allow the robot to reduce burden on the human supervisor by only requesting interventions when necessary, switching between robot control and human control based on some intervention criterion.~\citet{lazydagger} formalize the idea of supervisor burden as the expected total cost incurred by the human in providing interventions, which consists of the expected cost due to \textit{context switching} between autonomous and human control and the time spent actually providing interventions. However, it is difficult to design intervention criteria that limit this burden while ensuring that the robot gains sufficient information to imitate the supervisor's policy.

\begin{figure}
    \centering
    \includegraphics[width=0.9\textwidth]{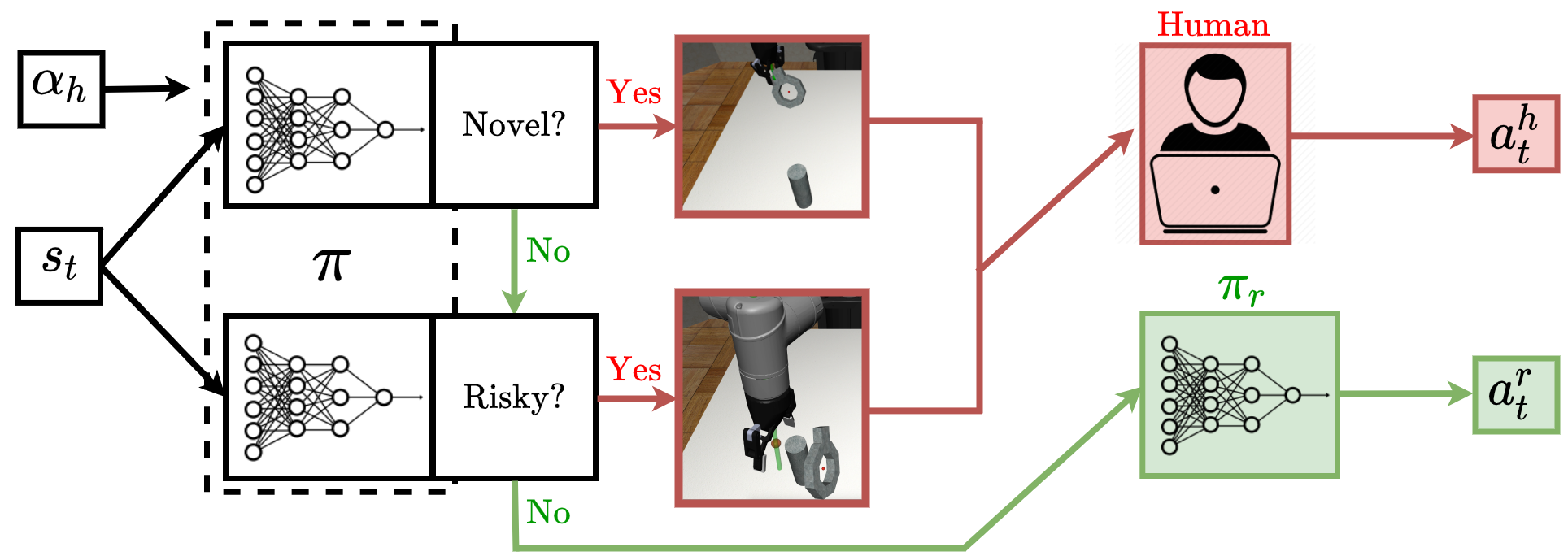}
    \caption{\textbf{\algname: }Given a desired context switching rate $\alphasupdesired$, \algname transfers control to a human supervisor if the current state $s_t$ is (1) sufficiently novel or (2) sufficiently risky, indicating that the probability of task success is low under robot policy $\pi_r$. Intuitively, one should not only distrust $\pi_r$ in states significantly out of the distribution of previously-encountered states, but should also cede control to a human supervisor in more familiar states where the robot predicts that it is unlikely to successfully complete the task.}
    \vspace{-0.2in}
    \label{fig:splash}
\end{figure}

% Our key insight and solution
% This paper makes the following contributions: (1) learned predictions of the novelty and risk of states in the environment, (2) \algname, an interactive imitation learning algorithm that maintains a desired level of human intervention requests, and (3) experiments demonstrating the effectiveness of \algname at reducing supervisor burden while learning challenging tasks in both simulation and on a complex physical cable routing task.

This paper makes several contributions. First, we develop intervention criteria based on a synthesis of two estimated properties of a given state: \textit{novelty}, which measures whether the state is significantly out of the distribution of previously encountered states, indicating that the robot policy should not be trusted; and \textit{risk}, which indicates whether the robot is unlikely to make task progress. \new{While state novelty has been considered in prior work~\cite{ensemble_dagger}, the key insight in our intervention criteria lies in combining novelty with a new risk metric to estimate the probability of task success.} Second, we present a new robot-gated interactive IL algorithm, \algnamefull (Figure~\ref{fig:splash}), which employs these measures jointly to solicit human interventions only when necessary.
Third, while prior robot-gated algorithms~\cite{safe_dagger, lazydagger} require careful parameter tuning to modulate the timing and frequency of human intervention requests, \algname only requires the supervisor to specify a desired context switching rate and \new{sets thresholds} accordingly.
% Additionally, \algname allows the supervisor to specify a target context switching rate and then automatically modulates other parameters to determine the timing and frequency of interventions; thus, \algname requires minimal parameter tuning compared to prior robot-gated algorithms~\cite{safe_dagger, lazydagger}. 
Fourth, experimental results demonstrate \algname's effectiveness for reducing supervisor burden while learning challenging tasks both in simulation and in an image-based cable routing task on a physical robot. 
%\algname gracefully scales to high-dimensional robotic control tasks, including a challenging manipulation task in simulation and a visuomotor cable routing task on a physical robot, while simultaneously increasing task performance and limiting supervisor burden.
Finally, the results of a human user study applying \algname to control a fleet of three simulated robots suggest that \algname significantly improves performance on both the robots' task and an independent human task while imposing fewer context switches, fewer human intervention actions, and lower mental load and frustration than prior algorithms.
\section{Related Work}
\label{sec:related-work}
\textbf{Imitation Learning from Human Feedback:}
% Briefly talk about issues with fully offline IL (covariate shift, etc...)
There has been significant prior work in offline imitation learning, in which the agent leverages an offline dataset of expert demonstrations either to directly match the distribution of trajectories in the offline dataset~\cite{pomerleau1991efficient, ho2016generative, argall2009survey, osa2018algorithmic, arora2018survey, ijspeert2013dynamical,paraschos2013probabilistic}, for instance via Behavior Cloning~\cite{torabi2018behavioral, bc_driving}, or to learn a reward function that can then be optimized via reinforcement learning~\cite{abbeel2004apprenticeship,ho2016generative,brown2019drex}. However, while these approaches have shown significant success in a number of domains~\cite{one-shot-visual-IL, rope-descriptors, ganapathi2020learning, bc_driving}, learning from purely offline data leads to a trajectory distribution mismatch which yields suboptimal performance both in theory and practice~\cite{dagger, DART}. To address this problem, there have been a number of approaches that utilize online human feedback while the agent acts in the environment, such as providing suggested actions~\cite{dagger, converging-supervisors,judah2011active,jauhri2020interactive} or preferences~\cite{sadigh2017active,christiano2017deep,ibarz2018reward,palan2019learning,biyik2019asking,brown2020safe}. However, many of these forms of human feedback may be unreliable if the robot visits states that significantly differ from those the human supervisor would themselves visit; in such situations, it is challenging for the supervisor to determine what correct behavior should look like without directly interacting with the environment~\cite{EIL,reddy2018shared}. 
% Furthermore, allowing only the robot to act in the environment can be unsafe, and most of these algorithms result in relatively high supervisor burden.
% DAgger~\cite{dagger} requires the supervisor to suggest an action at every timestep, and the preference learning approaches require the supervisor to indicate a preference over every pair of executed actions or trajectories. SHIV~\cite{SHIV} reduces the required number of suggested actions but still faces the aforementioned pitfalls.

\textbf{Interactive Imitation Learning: }
A natural way to collect reliable online feedback for imitation learning is to periodically cede control to a human supervisor, who then provides a corrective intervention to illustrate desired behavior.
% Human-gated interactive IL
Human-gated interactive IL algorithms~\cite{hg_dagger, EIL, HITL} such as HG-DAgger require the human to determine when to engage in interventions. However, these algorithms require a human to continuously monitor the robot to determine when to intervene, which imposes significant burden on the supervisor and is particularly impractical if a small number of humans must supervise a large number of robots. Furthermore, it requires the human to determine when the robot needs help and when to cede control, which can be unintuitive and unreliable.

% Robot-gated interactive IL
By contrast, robot-gated interactive IL algorithms, such as EnsembleDAgger~\cite{ensemble_dagger}, SafeDAgger~\cite{safe_dagger}, and LazyDAgger~\cite{lazydagger}, allow the robot to actively query for human interventions. In practice, these algorithms estimate various quantities correlated with task performance~\cite{safe_dagger, lazydagger, aggrevate, SHIV} and uncertainty~\cite{ensemble_dagger} and use them to determine when to solicit interventions. Prior work has proposed intervention criteria which use the novelty of states visited by the robot~\cite{ensemble_dagger} or the predicted discrepancy between the actions proposed by the robot policy and by the supervisor~\cite{safe_dagger, lazydagger}. However, while state novelty provides a valuable signal for soliciting interventions, we argue that this alone is insufficient, as a state's novelty does not convey information about the level of precision with which actions must be executed in that state. In practice, many robotic tasks involve moving through critical ``bottlenecks"~\cite{HITL}, which, though not necessarily novel, still present challenges. Examples include moving an eating utensil close to a person's mouth or placing an object on a shelf without disturbing nearby objects. Similarly, even if predicted accurately, action discrepancy is often a flawed risk measure, as high action discrepancy between the robot and the supervisor may be permissible when fine-grained control is not necessary (e.g. a robot gripper moving in free space) but impermissible when precision is critical (e.g. a robot gripper actively trying to grasp an object). In contrast, \algname presents an intervention criteria incorporating both state novelty and a novel risk metric and automatically tunes key parameters, allowing efficient use of human supervision.
% \algname also significantly reduces the need for parameter tuning, to which prior algorithms can be highly sensitive~\cite{ensemble_dagger}.
\section{Problem Statement}
\label{sec:prob-statement}
Given a robot, a task for the robot to accomplish, and a human supervisor with a specified context switching budget, the goal is to train the robot to imitate supervisor performance within the budget. We model the robot environment as a discrete-time Markov Decision Process (MDP) $\mathcal{M}$ with continuous states $s \in \mathcal{S}$, continuous actions $a \in \mathcal{A}$, and time horizon $T$ \cite{puterman2014markov}. We consider the interactive imitation learning (IL) setting~\cite{hg_dagger}, where the robot does not have access to a shaped reward function or to the MDP's transition dynamics but can temporarily cede control to a supervisor who uses policy $\pisup: \mathcal{S} \to \mathcal{A}$. We specifically focus on tasks where there is a goal set $\mathcal{G}$ which determines success, but that can be challenging and long-horizon, making direct application of RL highly sample inefficient.

We assume that the human and robot utilize the same action space (e.g. through a teleoperation interface) and that task success can be specified by convergence to some goal set $\mathcal{G} \subseteq \mathcal{S}$ within the time horizon (i.e., the task is successful if $\mathcal{G}$ is reached within $T$ timesteps). We further assume access to an indicator function $\mathds{1}_{\mathcal{G}}: \mathcal{S} \rightarrow \{0, 1\}$, which indicates whether a state belongs to the goal set $\mathcal{G}$.

The IL objective is to minimize a surrogate loss function $J(\pirob)$ to encourage the robot policy $\pirob: \mathcal{S} \to \mathcal{A}$ to match $\pisup$:
\begin{align}
    \label{eq:IL-objective}
    J(\pirob) = \sum_{t=1}^{T} \mathds{E}_{s_t \sim d^{\pirob}_t} \left[\mathcal{L}(\pirob(s_t),\pisup(s_t))\right],
\end{align}
where $\mathcal{L}(\pirob(s),\pisup(s))$ is an action discrepancy measure between $\pirob(s)$ and $\pisup(s)$ (e.g. MSE loss), and $d^{\pirob}_t$ is the marginal state distribution at timestep $t$ induced by the robot policy $\pirob$ in $\mathcal{M}$. 

In the interactive IL setting, meanwhile, in addition to optimizing Equation~\eqref{eq:IL-objective}, a key design goal is to minimize the imposed burden on the human supervisor. To formalize this, we define a switching policy $\pimeta$, which determines whether the system is under robot control $\pirob$ (which we call \textit{autonomous mode}) or human supervisor control $\pisup$ (which we call \textit{supervisor mode}). Following prior work~\cite{lazydagger}, we define $C(\pimeta)$, the expected number of \textit{context switches} in an episode under policy $\pimeta$, as follows: ${C(\pimeta) = \sum_{t=1}^{T} \mathds{E}_{s_t \sim d^{\pimeta}_t} \left[ \indicatorint(s_t; \pimeta) \right]}$, where $\indicatorint(s_t; \pimeta)$ is an indicator for whether or not a context switch occurs from autonomous to supervisor control. Similarly, we define $I(\pimeta)$ as the expected number of \textit{supervisor actions} in an intervention solicited by $\pimeta$.
% the expected number of \textit{supervisor actions} in an episode under policy $\pimeta$, as follows: ${D(\pimeta) = \sum_{t=1}^{T} \mathds{E}_{s_t \sim d^\pimeta_t} \left[ \indicatorsup(s_t; \pimeta) \right]}$, where $\indicatorsup(s_t; \pimeta)$ indicates whether or not the action at $s_t$ is provided by the supervisor.
We then define the total burden $B(\pimeta)$ imposed on the human supervisor as follows:
\begin{align}
    \label{eq:burden}
        B(\pimeta) = C(\pimeta)\cdot \big(L + I(\pimeta)\big),
\end{align}
where $L$ is the \textit{latency} of a context switch between control modes (summed over both switching directions) in units of timesteps (one action per timestep). The interactive IL objective is to minimize the discrepancy from the supervisor policy while limiting supervisor burden within some $\Gamma_{\rm b}$:
\begin{align}
\begin{split}
    \label{eq:LazyDAgger-objective}
    \pimeta &= \argmin_{\pimeta' \in \Pi}\{ J(\pirob)
    \mid B(\pimeta') \leq \Gamma_{\rm b}\}.
\end{split}
\end{align}
\new{Because it is challenging to explicitly optimize policies to satisfy the supervisor burden constraint in Equation~\eqref{eq:LazyDAgger-objective}, we present novel intervention criteria that enable reduction of supervisor burden by limiting the total number of interventions to a user-specified budget. Given sufficiently high latency $L$, limiting the interventions $C(\pi)$ directly corresponds to limiting supervisor burden $B(\pi)$.}
\section{\algnamefull}
\label{sec:alg}
\algname determines when to switch between autonomous and human supervisor control modes by leveraging estimates of both the \textit{novelty} and \textit{risk} of states. Below, Sections~\ref{subsec:novelty} and \ref{subsec:risk} discuss the estimation of state novelty and risk of task failure, respectively, while Section~\ref{subsec:switches} discusses \algname's integration of these measures to determine when to switch control modes. Section~\ref{subsec:tuning} then describes an \new{online procedure to set thresholds for switching between control modes}. Finally, Section~\ref{subsec:overview} describes the full control flow of \algname.

\subsection{Novelty Estimation}
\label{subsec:novelty}
When the robot policy visits states that lie significantly outside the distribution of those encountered in the supervisor trajectories, it does not have any reference behavior to imitate. This motivates initiating interventions to illustrate desired recovery behaviors in these states. However, estimating the support of the state distribution visited by the human supervisor is challenging in the high-dimensional state spaces common in robotics. Following prior work~\cite{ensemble_dagger}, we train an ensemble of policies with bootstrapped samples of transitions from supervisor trajectories. We then measure the novelty of a given state $s$ by calculating the variance of the policy outputs at state $s$ across ensemble members. In practice, the action $a \in \mathcal{A}$ outputted by each policy is a vector; thus, we measure state novelty by computing the variance of each component of the action vector $a$ across the ensemble members and then averaging over the components. We denote this quantity by $\textrm{Novelty}(s)$. Once in supervisor mode, as noted in~\citet{lazydagger}, we can obtain a more precise correlate of novelty by computing the ground truth action discrepancy between actions suggested by the supervisor and the robot policy.

\subsection{Risk Estimation}
\label{subsec:risk}
Interventions may be required not only in novel states outside the distribution of supervisor trajectories, but also in familiar states that are prone to result in task failure. \new{For example, a task might have a ``bottleneck" region with low tolerance for error, which has low novelty but nevertheless requires more supervision to learn a reliable robot policy.} To address this challenge, we propose a novel measure of a state's ``riskiness," capturing the likelihood that the robot cannot successfully converge to the goal set $\mathcal{G}$. We first define a Q-function to quantify the discounted probability of successful convergence to $\mathcal{G}$ from a given state and action under the robot policy:
\begin{align}
    \label{eq:RL}
    Q^{\pi_r}_{\mathcal{G}}(s_t, a_t) &= \mathbb{E}_{\pi_r}\left[ {\sum_{t'=t}^{\infty}\gamma^{t'-t} \mathds{1}_\mathcal{G}(s_t') | s_t, a_t} \right],
\end{align}
where $\mathds{1}_\mathcal{G}(s_t)$ is equal to 1 if $s_t$ belongs to $\mathcal{G}$. We estimate $Q^{\pi_r}_{\mathcal{G}}(s_t, a_t)$ via a function approximator $\safety$ parameterized by $\phi$, and define a state's riskiness in terms of this learned Q-function:
\begin{align}
    \label{eq:risk}
    \text{Risk}^{\pi_r}(s, a) = 1 - \safety(s, a).
\end{align}
In practice, we train $\safety$ on transitions $(s_t, a_t, s_{t+1})$ collected by the supervisor from both offline data and online interventions by minimizing the following MSE loss inspired by~\cite{recovery-rl}:
\begin{align}
    \label{eq:safetycritic}
    \begin{split}
    J^Q_{\mathcal{G}}(s_t, a_t, s_{t+1}; \phi) &= \frac{1}{2}\left (\safety(s_t, a_t) - (\mathds{1}_\mathcal{G}(s_t) + \right.
    \left. (1 - \mathds{1}_\mathcal{G}(s_t))\gamma \safety(s_{t+1}, \pi_r(s_{t+1})))\right)^2.
    \end{split}
\end{align}
% Note that since updates to $\pirob$ are infrequent and decoupled from updates to $\safety$ (Section~\ref{subsec:overview}), we require fewer samples than standard reinforcement learning to train an accurate critic. 
Note that since $\safety$ is only used to solicit interventions, it must only be accurate enough to distinguish risky states from others, rather than be able to make the fine-grained distinctions between different states required for accurate policy learning in reinforcement learning.
%Inspired by prior work~\cite{recovery-rl}, we leverage the risk measure in Equation~\eqref{eq:risk} to define a switching controller as discussed below in Section~\ref{subsec:switches}.

\subsection{Regulating Switches in Control Modes}
\label{subsec:switches}
We now describe how \algname leverages the novelty estimator from Section~\ref{subsec:novelty} and the risk estimator from Section~\ref{subsec:risk} to regulate switches between autonomous and supervisor control. While in autonomous mode, the switching policy $\pimeta$ initiates a switch to supervisor mode at timestep $t$ if either (1) state $s_t$ is sufficiently unfamiliar or (2) the robot policy has a low probability of task success from $s_t$. Stated precisely, $\pimeta$ initiates a switch to supervisor mode from autonomous mode at timestep $t$ if the predicate $\textrm{Intervene}(s_t, \deltasup, \tausup)$ evaluates to $\textsc{True}$, where $\textrm{Intervene}(s_t, \deltasup, \tausup)$ is $\textsc{True}$ if (1) $\textrm{Novelty}(s_t) > \deltasup$ or (2) %$\textrm{Novelty}(s_t) \leq \delta \textsc{ and } 
$\text{Risk}^{\pi_r}(s_t, \pi_r(s_t)) > \tausup$ and $\textsc{False}$ otherwise. Note that the proposed switching policy only depends on $\text{Risk}^{\pi_r}$ for states which are \textit{not} novel (as novel states already initiate switches to supervisor control regardless of risk), since the learned risk measure should only be trusted on states in the neighborhood of those on which it has been trained.

In supervisor mode, $\pimeta$ switches to autonomous mode if the action discrepancy between the human and robot policy and the robot's task failure risk are both below threshold values (Section~\ref{subsec:tuning}), indicating that the robot is in a familiar and safe region. Stated precisely, $\pimeta$ switches to autonomous mode from supervisor mode if the predicate $\textrm{Cede}(s_t, \deltaauto, \tauauto)$ evaluates to $\textsc{True}$, where $\textrm{Cede}(s_t, \deltaauto, \tauauto)$ is $\textsc{True}$ if (1) $||\pirob(s_t) - \pisup(s_t)||_2^2 < \deltaauto$ and (2) $\text{Risk}^{\pi_r}(s_t, \pi_r(s_t)) < \tauauto$, and $\textsc{False}$ otherwise. \new{Here, the risk metric ensures that the robot has a high probability of autonomously completing the task, while the coarser 1-step action discrepancy metric verifies that we are in a familiar region of the state space where the $\safety$ values can be trusted.} Motivated by prior work~\cite{lazydagger} and hysteresis control~\cite{hysteresis}, we use stricter switching criteria in supervisor mode ($\tauauto < \tausup$) to encourage lengthier interventions and reduce context switches experienced by the human supervisor.

\subsection{\new{Computing Risk and Novelty Thresholds from Data}}
\label{subsec:tuning}
One challenge of the control strategy presented in Section~\ref{subsec:switches} lies in tuning the key parameters ($\deltasup, \deltaauto, \tausup, \tauauto$) governing when context switching occurs. As noted in prior work~\cite{ensemble_dagger}, performance and supervisor burden can be sensitive to these thresholds. To address this difficulty, we assume that the user specifies their availability in the form of a desired intervention budget $\alphasupdesired \in [0, 1]$, indicating the desired proportion of timesteps in which interventions will be requested. This desired context switching rate can be interpreted in the context of supervisor burden as defined in Equation~\eqref{eq:burden}: if the latency of a context switch dominates the time cost of the intervention itself, limiting the expected number of context switches to within some intervention budget directly limits supervisor burden.

Given $\alphasupdesired$, we set $\tausup$ to be the $(1-\alphasupdesired)$-quantile of $\text{Risk}^{\pi_r}(s, \pi_r(s))$ for all states previously visited by $\pirob$ and set $\deltasup$ to be the $(1-\alphasupdesired)$-quantile of $\textrm{Novelty}(s)$ for all states previously visited by $\pirob$. We set $\deltaauto$ to be the mean action discrepancy on the states visited by the supervisor after $\pirob$ is trained and set $\tauauto$ to be the median of $\text{Risk}^{\pi_r}(s, \pi_r(s))$ for all states previously visited by $\pirob$. (Note that $\tauauto$ can easily be set to different quantiles to adjust mean intervention length if desired.) We find that these settings strike a balance between informative interventions and imposed supervisor burden.

\subsection{\algname Overview}
\label{subsec:overview}
%The full \algname procedure is summarized in Algorithm~\ref{alg:main}.
We now summarize the \algname procedure, with full pseudocode available in the supplement. \algname first initializes $\pi_r$ via Behavior Cloning on offline transitions $(\mathcal{D}_h$ from the human supervisor, $\pi_h$). Then, $\pi_r$ collects an initial offline dataset $\mathcal{D}_r$ from the resulting $\pi_r$, initializes $\safety$ by optimizing Equation~\eqref{eq:risk} on $\mathcal{D}_r \cup \mathcal{D}_h$, and initializes parameters $\tausup, \tauauto, \deltasup$, and $\deltaauto$ as in Section~\ref{subsec:tuning}. We then collect data for $N$ episodes, each with up to $T$ timesteps. In each timestep of each episode, we determine whether robot policy $\pi_r$ or human supervisor $\pi_h$ should be in control using the procedure in Section~\ref{subsec:switches}. Transitions in autonomous mode are aggregated into $\mathcal{D}_r$  while transitions in supervisor mode are aggregated into $\mathcal{D}_h$. After each episode, $\pi_r$ is updated via supervised learning on $\mathcal{D}_h$, while $\safety$ is updated on $\mathcal{D}_r \cup \mathcal{D}_h$ to reflect the probability of task success of the resulting $\pi_r$.

\section{Experiments}
\label{sec:exps}
In the following experiments, we study whether \algname can balance task performance and supervisor burden more effectively than prior IL algorithms in three contexts: (1) training a simulated robot to perform a peg insertion task (Section~\ref{subsec:sim}); (2) supervising a fleet of three simulated robots to perform the peg insertion task in a human user study (Section~\ref{subsec:fleet}); and (3) training a physical surgical robot to perform a cable routing task (Section~\ref{subsec:physical}). \new{In the supplementary material, we also include results from an additional simulation experiment on a challenging block stacking task.}
\subsection{Evaluation Metrics}
%\algname has two primary use cases: (1) leveraging interventions during training to maximize policy performance at execution time and (2) improving policy performance by leveraging access to a human-in-the-loop. These two cases motivate two methods of evaluating learned policies during execution after training has completed:

We consider \algname's performance during training and execution. For the latter, we evaluate both the (1) \textit{autonomous success rate}, or success rate when deployed after training without access to a human supervisor, and (2) \textit{intervention-aided success rate}, or success rate when deployed after training with a human supervisor in the loop. These metrics are reported in the Peg Insertion study (Section~\ref{subsec:sim}) and the Physical Cable Routing study (Section~\ref{subsec:physical}). For all experiments, during both training and intervention-aided execution, we evaluate the number of interventions, human actions, and robot actions per episode. These metrics are computed over successful episodes only to prevent biasing the metrics by the maximum episode horizon length $T$; \new{such bias occurs, for instance, when less successful policies appear to take more actions due to hitting the time boundary more often. Additional metrics including cumulative statistics across all episodes are reported in the supplement.}
%In the Peg Insertion simulation experiment and Physical Cable Routing experiment, we further report the percentage of training time-steps that involve a context switch.
%We additionally evaluate \algname in a user study with three robots performing the simulated peg insertion task (Section~\ref{subsec:fleet}), in which participants attempt to maximize throughput (total number of task successes across the three robots within a budget of total allowed actions) during training while context switching between helping the robots and performing a distractor Concentration game task. 
In our user study (Section~\ref{subsec:fleet}), we also report the following quantities: throughput (total number of task successes across the three robots), performance on an independent human task, the idle time of the robots in the fleet, and users' qualitative ratings of mental load and frustration. \new{By comparing the amount of human supervision and success rates across different algorithms, we are interested in evaluating how effectively each algorithm balances supervision with policy performance.}

\subsection{Comparisons}
\label{subsec:comparisons}
We compare \algname to the following algorithms: Behavior Cloning, which does not use interventions;
%and is trained via supervised learning on offline demonstrations
HG-DAgger~\cite{hg_dagger}, which is human-gated and always requires supervision; SafeDAgger~\cite{safe_dagger}, which is robot-gated and performs interventions based on estimated action discrepancy between the human supervisor and robot policy; and LazyDAgger~\cite{lazydagger}, which builds on SafeDAgger by introducing an asymmetric switching criterion to encourage lengthier interventions. We also implement two ablations: one that does not use a novelty measure to regulate context switches (\algname(-Novelty)) and one that does not use risk to regulate context switches (\algname(-Risk)).
% These baseline algorithms are described below:
% \begin{itemize}[topsep=0pt, leftmargin=*]
%     \item \textit{Behavior Cloning: } The robot policy is trained entirely from offline task demonstrations. We calibrate the number of offline transitions to be slightly greater than the total number of human actions seen by \algname.
%     \item \textit{HG-DAgger~\cite{hg_dagger}: } Human-gated interactive IL baseline where the human monitors task execution and intervenes when deemed necessary (e.g. lack of task progress).
%     \item \textit{SafeDAgger~\cite{safe_dagger}: } Robot-gated interactive IL algorithm which uses the estimated action discrepancy between the robot and human policy to determine when to solicit interventions.
%     \item \textit{LazyDAgger~\cite{lazydagger}: } Robot-gated interactive IL algorithm which builds on~\cite{safe_dagger} but encourages fewer context switches via asymmetric switching criteria.
%     \item \textit{Ours(-Risk):} \algname, but without a learned risk measure. Here, we switch from autonomous to supervisor mode based only on measured novelty and return to autonomous mode if the action discrepancy between the robot policy and human action is sufficiently low. This corresponds to the ``doubt rule" from EnsembleDAgger~\cite{ensemble_dagger} combined with the action discrepancy metric from~\cite{lazydagger} and the automatic parameter tuning from \algname.
%     \item \textit{Ours(-Novelty):} \algname, but does not use any novelty measure to determine when to solicit interventions.
% \end{itemize}

\subsection{Peg Insertion in Simulation}
\label{subsec:sim}
\begin{figure*}
    \centering
    \includegraphics[width=0.9\textwidth]{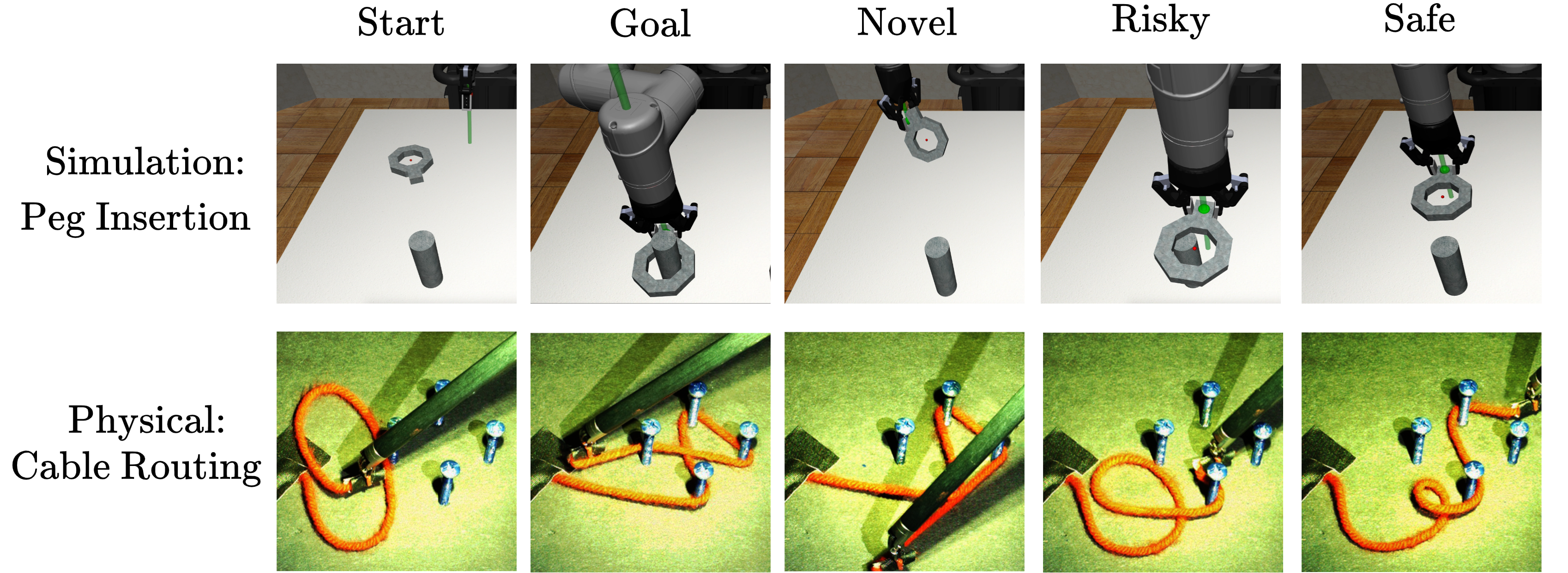}
    \caption{\textbf{Experimental Domains: }We visualize the peg insertion simulation domain (top row) and the physical cable routing domain with the physical robot (bottom row). We visualize sample start and goal states, in addition to states which \algname categorizes as novel, risky, and neither. \algname marks states as novel if they are far from behavior that the supervisor would produce, and risky if it is stuck in a bottleneck, e.g. if the ring is wedged against the side of the cylinder (top) or the cable is near all four obstacles (bottom).}
    \label{fig:vis}
    \vspace{-0.2in}
\end{figure*}

We first evaluate \algname on a long-horizon peg insertion task (Figure~\ref{fig:vis}) from the Robosuite simulation environment~\cite{robosuite}. The goal is to grasp a ring in a random initial pose and thread it over a cylinder at a fixed target location. This task has two bottlenecks which motivate learning from interventions: (1) correctly grasping the ring and (2) correctly placing it over the cylinder (Figure~\ref{fig:vis}). A human teleoperates the robot through a keyboard interface to provide interventions. The states consist of the robot's joint angles and ring's pose, while the actions specify 3D translation, 3D rotation, and opening or closing the gripper.
%on the end effector.
For \algname and its ablations, we use target intervention frequency $\alphasupdesired=0.01$ and set other parameters via the automated tuning method (Section~\ref{subsec:tuning}). We collect 30 offline task demos (2,687 state-action pairs) from a human supervisor to initialize the robot policy for all compared algorithms. Behavior Cloning is given \new{additional state-action pairs roughly equivalent to the average amount of supervisor actions solicited by the interactive algorithms (Table~\ref{tab:insertion-extra} in the appendix)}. For \algname and each interactive IL baseline, we perform 10,000 environment steps, during which each episode takes at most 175 timesteps and system control switches between the human and robot. Hyperparameter settings for all algorithms are detailed in the supplement.

Results (Table~\ref{tab:sim}) suggest that \algname achieves a significantly higher autonomous success rate than prior robot-gated algorithms, although it does request more human actions due to its conservative exit criterion for interventions ($\textrm{Cede}(s_t, \deltaauto, \tauauto)$). However, the number of interventions is similar to prior robot-gated algorithms, indicating that while \algname requires more human actions, it imposes a similar supervisor burden to SafeDAgger and LazyDAgger in settings in which context switches are expensive or time-consuming (e.g. high latency $L$ in Equation~\ref{eq:burden}). We find that all interactive IL algorithms substantially outperform Behavior Cloning, which does not have access to supervisor interventions. Notably, \algname achieves a higher autonomous success rate than even HG-DAgger, in which the supervisor is able to decide the timing and length of interventions. This indicates that \algname's intervention criteria enable it to autonomously solicit interventions \new{as informative as those chosen by a human supervisor with expert knowledge of the task}. Furthermore, \algname achieves a 100\% intervention-aided success rate at execution time, suggesting that \algname successfully identifies the required states at which to solicit interventions.  We find that both ablations of \algname (Ours (-Novelty) and Ours (-Risk)) achieve significantly lower autonomous success rates, indicating that both the novelty and risk measures are critical to \algname's performance. We calculate \algname's context switching rate to be 1.15\% novelty switches and 0.79\% risk switches, both approximately within the budget of $\alphasupdesired=0.01$.

\begin{table}[t]
\caption{\textbf{Peg Insertion in Simulation Results: }We first report training performance (number of interventions (Ints), number of human actions (Acts (H)), and number of robot actions (Acts (R))) and report the success rate of the fully-trained policy at execution time when no interventions are allowed (Auto Succ.). We then evaluate the fully-trained policies with interventions allowed and report the same intervention statistics and the success rate (Int-Aided Succ.). We find that \algname achieves the highest autonomous and intervention-aided success rates among all algorithms compared. Notably, \algname even achieves a higher autonomous success rate than HG-DAgger, in which the human decides when to intervene during training.}
\centering
\resizebox{\columnwidth}{!}{%
\begin{tabular}{|l||r|r|r||r||r|r|r||r||}
    \hline
    \multicolumn{1}{|c||}{Algorithm} &
    \multicolumn{3}{|c||}{Training Interventions} &
    \multicolumn{1}{|c||}{Auto Succ.} & 
    \multicolumn{3}{|c||}{Execution Interventions} & \multicolumn{1}{|c||}{Int-Aided Succ.} \\
    \cline{2-4}
    \cline{6-8}
      \multicolumn{1}{|c||}{} &
      \multicolumn{1}{|c|}{Ints} &
      \multicolumn{1}{|c|}{Acts (H)} &
      \multicolumn{1}{|c||}{Acts (R)} &
      \multicolumn{1}{|c||}{} &
      \multicolumn{1}{|c|}{Ints} &
      \multicolumn{1}{|c|}{Acts (H)} &
      \multicolumn{1}{|c||}{Acts (R)} &
      \multicolumn{1}{|c||}{}\\
    \hline
    \cline{1-9}
     Behavior Cloning & N/A & N/A & $108.0 \pm 15.9$ & $24/100$ & N/A & N/A & N/A & N/A \\ 
     SafeDAgger & $3.89 \pm 1.44$ & $19.8 \pm 9.9$ & $88.8 \pm 19.4$  & $24/100$ & $4.00 \pm 1.37$ & $19.5 \pm 5.3$ & $77.5 \pm 11.7$ & $17/20$\\ 
     LazyDAgger & $1.46 \pm 1.15$ & $13.2 \pm 12.4$ & $102.1 \pm 18.2$ & $48/100$ & $1.73 \pm 1.29$ & $12.6 \pm 14.4$ & $91.7 \pm 24.0$ & $11/20$\\ 
     HG-DAgger & $1.49 \pm 0.88$ & $20.3 \pm 15.6$ & $97.1 \pm 17.5$ & $57/100$ & $1.15 \pm 0.73$ & $17.1 \pm 11.6$ & $103.6 \pm 14.0$ & $\mathbf{20/20}$\\ 
     Ours (-Novelty) & $\mathbf{0.79 \pm 0.81}$ & $35.1 \pm 23.1$ & $70.0 \pm 35.8$ & $49/100$ & $\mathbf{0.33 \pm 0.62}$ & $2.5 \pm 5.0$ & $114.0 \pm 26.0$  & $12/20$\\ 
     Ours (-Risk)  & $0.99 \pm 0.96$ & $7.8 \pm 12.0$ & $104.2 \pm 19.2$  & $49/100$ & $1.39 \pm 0.95$ & $9.8 \pm 12.0$ & $109.1 \pm 22.9$& $18/20$ \\ 
     Ours: \algname & $0.88 \pm 1.01$ & $43.6 \pm 24.5$ & $60.0 \pm 32.8$ & $\mathbf{73/100}$ & $1.35 \pm 0.66$ & $21.3 \pm 15.0$ & $84.8 \pm 21.8$  & $\mathbf{20/20}$ \\
    \hline
\end{tabular}}
\label{tab:sim}
\vspace{-0.2in}
\end{table}

\subsection{User Study: Controlling A Fleet of Three Robots in Simulation}
\label{subsec:fleet}
We conduct a user study with 10 participants (7 male and 3 female, aged 18-37). Participants supervise a fleet of three simulated robots, each performing the peg insertion task from Section~\ref{subsec:sim}. We evaluate how different interactive IL algorithms affect the participants' (1) ability to provide effective robot interventions, (2) performance on a distractor task performed between robot interventions, and (3) levels of mental demand and frustration.
% First, participants are given a few minutes to gain familiarity with the teleoperation interface in the Robosuite simulation environment~\cite{robosuite} and with the distractor task. 
For the distractor task, we use the game Concentration (also known as Memory or Matching Pairs), in which participants identify as many pairs of matching cards as possible among a set of face-down cards. This is intended to emulate tasks which require continual focus, such as cooking a meal or writing a research paper, in which frequent context switches between performing the task and helping the robots is frustrating and degrades performance.

The participants teleoperate the robots using three robot-gated interactive IL algorithms: SafeDAgger, LazyDAgger, and \algname. The participant is instructed to make progress on the distractor task only when no robot requests an intervention. When an intervention is requested, the participant is instructed to pause the distractor task, provide an intervention from the requested state until the robot (or multiple robots queued after each other) no longer requires assistance, and then return to the distractor task. The participants also teleoperate with HG-DAgger, where they no longer perform the distractor task and are instructed to continually monitor all three robots simultaneously and decide on the length and timing of interventions themselves. 
% The order of the algorithms is randomized for each participant to control for improvement in the distractor task and/or teleoperation over time.
Each algorithm runs for 350 timesteps, where in each timestep, all robots in autonomous mode execute one action and the human executes one action on the currently-supervised robot (if applicable). The supplement illustrates the user study interface and fully details the experiment protocol. All algorithms are initialized as in Section~\ref{subsec:sim}.

Results (Table ~\ref{tab:ustudy}) suggest that \algname achieves significantly higher throughput than all prior algorithms while requiring fewer interventions and fewer human actions, indicating that \algname requests interventions more judiciously than prior algorithms. Furthermore, \algname also enables a lower mean idle time for robots and higher performance on the distractor task. Notably, \algname solicits fewer interventions and total actions while achieving a higher throughput than HG-DAgger, in which the participant chooses when to intervene. %This suggests that when controlling a robot fleet, participants struggle to efficiently monitor the robots and make appropriate decisions on when to intervene, while \algname solicits interventions judiciously to achieve high throughput while still imposing low supervisor burden. 
We also report metrics of users' mental workload and frustration using the NASA-TLX scale~\cite{hart2006nasa} in the supplement. Results suggest that users experience lower degrees of frustration and mental load when interacting with \algname and LazyDAgger compared to HG-DAgger and SafeDAgger. We hypothesize that participants struggle with HG-DAgger due to the difficultly of monitoring multiple robots simultaneously, while SafeDAgger's frequent context switches lead to user frustration during experiments.

% \begin{figure*}
%     \centering
%     \includegraphics[width=\textwidth]{figures/user_study_survey.png}
%     \caption{\textbf{User Study Survey Results: }We illustrate the user study interface for the human-gated and robot-gated algorithms (left) and users' survey responses regarding their mental load and frustration (right) for each algorithm. Results suggest that users report similar levels of mental load and frustration for \algname and LazyDAgger, but significantly higher levels of both metrics for HG-DAgger and SafeDAgger. We hypothesize that the sparing and sustained interventions solicited by \algname and LazyDAgger lead to greater user satisfaction and comfort compared to algorithms which force the user to constantly monitor the system (HG-DAgger) or frequently context switch between teleoperation and the distractor task.}
%     \label{fig:survey_results}
% \end{figure*}

\begin{table}[t]
\caption{\textbf{Three-Robot Fleet Control User Study Results: } Results for experiments with 10 human subjects and 3 simulated robots on the peg insertion task. We report the total numbers of interventions, human actions, and robot actions, as well as the throughput, or total task successes achieved across robots, for all algorithms. Additionally, for robot-gated algorithms, we report the Concentration score (number of pairs found) and the mean idle time of robots in the fleet in timesteps. Results suggest that \algname outperforms all prior algorithms across all metrics, requesting fewer interventions and total human actions while achieving higher throughput, lowering the robots' mean idle time, and enabling higher performance on the Concentration task. %This suggests that \algname's intervention criteria significantly improve robot and human productivity.
}
\centering
\resizebox{\columnwidth}{!}{%
\begin{tabular}{l | r r r r r r}
    Algorithm & Interventions & Human Actions & Robot Actions & Concentration Pairs & Throughput & Mean Idle Time \\ \hline
    HG-DAgger & 10.6 $\pm$ 2.5 & 198.0 $\pm$ 32.1 & 834.4 $\pm$ 38.1 & N/A & 5.1 $\pm$ 1.9 & N/A\\
    SafeDAgger & 22.1 $\pm$ 4.8 & 234.1 $\pm$ 31.8 & 700.7 $\pm$ 70.4 & 17.7 $\pm$ 8.2 & 3.0 $\pm$ 2.4 & 38.4 $\pm$ 14.1 \\ 
    LazyDAgger & 10.0 $\pm$ 2.1 & 219.5 $\pm$ 43.3 & 719.2 $\pm$ 89.7 & 20.9 $\pm$ 7.9 & 5.1 $\pm$ 1.7 & 37.1 $\pm$ 20.5 \\ 
    Ours: \algname & $\mathbf{7.9 \pm 2.1}$ & $\mathbf{179.4 \pm 34.9}$ & 793.2 $\pm$ 86.6 & $\mathbf{33.0 \pm 8.5}$ & $\mathbf{9.2 \pm 2.0}$ & $\mathbf{25.8 \pm 19.3}$ \\ 
\end{tabular}}
\label{tab:ustudy}
\vspace{-0.1in}
\end{table}

\subsection{Physical Experiment: Visuomotor Cable Routing}
\label{subsec:physical}
Finally, we evaluate \algname on a long-horizon cable routing task with a da Vinci surgical robot~\cite{kazanzides2014open}. Here, the objective is to route a red cable into a Figure-8 pattern around 4 pegs via teleoperation with the robot's master controllers (see supplement). The algorithm only observes high-dimensional $64 \times 64 \times 3$ RGB images of the workspace and generates continuous actions representing delta-positions in $(x, y)$. As in Section~\ref{subsec:sim}, \algname uses a target intervention frequency of $\alphasupdesired=0.01$. We collect 25 offline task demonstrations (1,381 state-action pairs) from a human supervisor to initialize the robot policy for \algname and all comparisons. We perform 1,500  environment steps, where each episode has at most 100 timesteps and system control can switch between the human and robot. The supplement details the hyperparameter settings for all algorithms.

Results (Table~\ref{tab:physical}) suggest that both \algname and HG-DAgger achieve a significantly higher autonomous success rate than Behavior Cloning, which is never able to complete the task. Furthermore, \algname achieves a \new{comparable} autonomous success rate to HG-DAgger while requesting fewer interventions and a similar number of total human actions. This again suggests that \algname's intervention criteria enable it to solicit interventions \new{equally as informative or more informative} than those chosen by a human supervisor. Finally, at execution time \algname achieves a 100\% intervention-aided success rate with minimal supervision, again indicating that \algname successfully identifies the \new{timing and length of interventions} to increase policy reliability.

\begin{table}[t]
\caption{\textbf{Physical Cable Routing Results: }We first report intervention statistics during training (number of interventions (Ints), number of human actions (Acts (H)), and number of robot actions (Acts (R))) and report the success rate of the fully-trained policy at execution time when no interventions are allowed (Auto Succ.). We then evaluate the fully-trained policies with interventions allowed and report the same intervention statistics and the success rate (Int-Aided Succ.). We find that \algname achieves the highest autonomous and intervention-aided success rates among all algorithms compared. Notably, \algname even achieves a higher autonomous success rate than HG-DAgger, in which the human decides when to intervene during training. %We find that \algname achieves the highest autonomous and intervention-aided success rates among all algorithms compared. As in Table~\ref{tab:sim}, \algname achieves a higher autonomous success rate than HG-DAgger, in which the human decides when to intervene during training.
}
\centering
\resizebox{\columnwidth}{!}{%
\begin{tabular}{|l||r|r|r||r||r|r|r||r||}
    \hline
    \multicolumn{1}{|c||}{Algorithm} &
    \multicolumn{3}{|c||}{Training Interventions} &
    \multicolumn{1}{|c||}{Auto Succ.} & 
    \multicolumn{3}{|c||}{Execution Interventions} & \multicolumn{1}{|c||}{Int-Aided Succ.}\\
    \cline{2-4}
    \cline{6-8}
      \multicolumn{1}{|c||}{} &
      \multicolumn{1}{|c|}{Ints} &
      \multicolumn{1}{|c|}{Acts (H)} &
      \multicolumn{1}{|c||}{Acts (R)} &
      \multicolumn{1}{|c||}{} &
      \multicolumn{1}{|c|}{Ints} &
      \multicolumn{1}{|c|}{Acts (H)} &
      \multicolumn{1}{|c||}{Acts (R)} &
      \multicolumn{1}{|c||}{}\\
    \hline
    \cline{1-9}
     Behavior Cloning & N/A & N/A & N/A & $0/15$ & N/A & N/A & N/A & N/A\\ 
     HG-DAgger & $1.55 \pm 1.16$ & $13.9 \pm 10.9$  & $55.5 \pm 10.9$  & $10/15$ & $\mathbf{0.40 \pm 0.49}$ & $2.7 \pm 3.5$ & $73.9 \pm 7.9$ & $\mathbf{15/15}$\\
     Ours: \algname & $\mathbf{1.42 \pm 1.14}$ & $15.2 \pm 12.4$  & $45.5 \pm 18.3$  & $\mathbf{12/15}$ & $0.40 \pm 0.71$ & $1.5 \pm 3.1$ & $61.3 \pm 6.5$ & $\mathbf{15/15}$\\ 
    \hline
\end{tabular}}
\label{tab:physical}
\vspace{-0.2in}
\end{table}
%\vspace{-0.18in}
\section{Discussion and Future Work}
\label{sec:discussion}
We present \algname, a scalable robot-gated interactive imitation learning algorithm that leverages learned estimates of state novelty and risk of task failure to reduce burden on a human supervisor during training and execution. Experiments suggest that \algname effectively enables long-horizon robotic manipulation tasks in simulation, on a physical robot, and for a three-robot fleet while limiting burden on a human supervisor.
%and automatically tuning key algorithmic parameters. We additionally evaluate \algname in a user study focused on controlling a fleet of three simulated robots and find that \algname enables higher throughput while simultaneously facilitating user progress on a distractor task.
In future work, we hope to apply ideas from \algname to interactive reinforcement learning and larger scale fleets of physical robots. We also hope to study how \algname's performance varies with the target supervisor burden (specified via $\alphasupdesired$). In practice, $\alphasupdesired$ could even be time-varying: for instance, $\alphasupdesired$ may be significantly lower at night, when human operators may have limited availability. Similarly, $\alphasupdesired$ may be set to a higher value during training than at deployment, when the robot policy is typically higher quality. 

% hope to cast \algname in the context of goal-conditioned reinforcement learning, where the learned risk measure is leveraged to update the robot policy to potentially outperform the demonstrator by explicitly minimizing the likelihood of task failure. We will also evaluate \algname in fleet learning experiments with a large number of simulated and physical robots learning different tasks. In addition, we plan to experimentally evaluate how \algname's performance varies with the target supervisor burden (specified via $\alphasupdesired$). In practice, $\alphasupdesired$ could even be time-varying: for instance, $\alphasupdesired$ may be significantly lower at night, when human operators may have limited availability. Similarly, $\alphasupdesired$ may be set to a higher value during training than at deployment, when the robot policy is typically higher quality. 

% The maximum paper length is 8 pages excluding references and acknowledgements, and 10 pages including references and acknowledgements

\clearpage
% The acknowledgments are automatically included only in the final version of the paper.
\acknowledgments{This research was performed at the AUTOLAB at UC Berkeley in affiliation with the Berkeley AI Research (BAIR) Lab and the CITRIS ``People and Robots" (CPAR) Initiative. The authors were supported in part by the Scalable Collaborative Human-Robot Learning (SCHooL) Project, NSF National Robotics Initiative Award 1734633, and by donations from Google, Siemens, Amazon Robotics, Toyota Research Institute, Autodesk, Honda, Intel, and Hewlett-Packard and by equipment grants from PhotoNeo, NVidia, and Intuitive Surgical. Any opinions, findings, and conclusions or recommendations expressed in this material are those of the author(s) and do not necessarily reflect the views of the sponsors. We thank our colleagues who provided helpful feedback, code, and suggestions, especially Vincent Lim and Zaynah Javed.}

%===============================================================================

% no \bibliographystyle is required, since the corl style is automatically used.
% \clearpage
% \begin{small}
\bibliography{corl}  % .bib
% \end{small}
\clearpage
% \newpage
\section{Appendix}
In Appendix~\ref{subsec:app_alg_details}, we discuss algorithmic details for \algname and all comparisons. Then, Appendix~\ref{subsec:app_impl_details} discusses implementation and hyperparameter details for all algorithms. In Appendix~\ref{subsec:app_exp_details}, we provide additional details about the simulation and physical experiment domains, and
in Appendix~\ref{subsec:app_user_study}, we describe the protocol and detailed results from the conducted user study.

\subsection{Algorithm Details}
% Explain how each algorithm works in detail
\label{subsec:app_alg_details}
% Detailed description of all algorithms, including ThriftyDAgger
Here we provide a detailed algorithmic description of \algname and all comparisons.

\subsubsection{\algname}
The full pseudocode for \algname is provided in Algorithm~\ref{alg:main}. \algname first initializes $\pi_r$ via Behavior Cloning on offline transitions $(\mathcal{D}_h$ from the human supervisor, $\pi_h$) (line 1-2). Then, $\pi_r$ collects an initial offline dataset $\mathcal{D}_r$ from the resulting $\pi_r$, initializes $\safety$ by optimizing Equation~\eqref{eq:risk} on $\mathcal{D}_r \cup \mathcal{D}_h$, and initializes parameters $\tausup, \tauauto, \deltasup$, and $\deltaauto$ as in Section~\ref{subsec:tuning} (lines 3-5). We then collect data for $N$ episodes, each with up to $T$ timesteps. In each timestep of each episode, we determine whether robot policy $\pi_r$ or human supervisor $\pi_h$ should be in control using the procedure in Section~\ref{subsec:switches} (lines 10-20). Transitions in autonomous mode are aggregated into $\mathcal{D}_r$  while transitions in supervisor mode are aggregated into $\mathcal{D}_h$. Episodes are terminated either when the robot reaches a valid goal state or has exhausted the time horizon $T$. At this point, we re-initialize the policy to autonomous mode and update parameters $\tausup, \tauauto, \deltasup$, and $\deltaauto$  as in Section~\ref{subsec:tuning} (lines 21-23). After each episode, $\pi_r$ is updated via supervised learning on $\mathcal{D}_h$, while $\safety$ is updated on $\mathcal{D}_r \cup \mathcal{D}_h$ to reflect the task success probability of the resulting $\pi_r$ (lines 24-26).

\subsubsection{Behavior Cloning}
We train policy $\pirob$ via direct supervised learning with a mean-squared loss to predict reference control actions given a dataset of (state, action) tuples. Behavior Cloning is trained only on full expert demonstrations collected offline from $\pisup$ and is not allowed access to online interventions. Thus, Behavior Cloning is trained only on dataset $\mathcal{D}_h$ (line 1, Algorithm~\ref{alg:main}) and the policy is frozen thereafter.
In our simulation experiments, Behavior Cloning is given 50\% more offline data than the other algorithms for a more fair comparison, such that the amount of additional offline data is approximately equal to the average amount of online data provided to the other algorithms.

\subsubsection{SafeDAgger}
SafeDAgger~\cite{safe_dagger} is an interactive imitation learning algorithm which selects between autonomous and supervisor mode using a classifier $f$ that discriminates between ``safe" states, for which $\pirob$'s proposed action is within some threshold $\tausup$ of that proposed by supervisor policy $\pisup$, and ``unsafe" states, for which this action discrepancy exceeds $\tausup$. SafeDAgger learns this classifier using dataset $\mathcal{D}_h$ from Algorithm~\ref{alg:main}, and updates $f$ online as $\mathcal{D}_h$ is expanded through human interventions. During policy rollouts, if $f$ marks a state as safe, the robot policy is executed (autonomous mode), while if $f$ marks a state as unsafe, the supervisor is queried for an action. While this approach can be effective in some domains~\cite{safe_dagger}, prior work~\cite{lazydagger} suggests that this intervention criterion can lead to excessive context switches between the robot and supervisor, and thus impose significant burden on a human supervisor. As in \algname and other DAgger~\cite{dagger} variants, SafeDAgger updates $\pirob$ on an aggregated dataset of all transitions collected by the supervisor (analogous to $\mathcal{D}_h$ in Algorithm~\ref{alg:main}).

\subsubsection{LazyDAgger}
LazyDAgger~\cite{lazydagger} builds on SafeDAgger~\cite{safe_dagger} and trains the same action discrepancy classifier $f$ to determine whether the robot and supervisor policies will significantly diverge at a given state. However, LazyDAgger introduces a few modifications to SafeDAgger which lead to lengthier and more informative interventions in practice. First, LazyDAgger observes that when the supervisor has control of the system (supervisor mode), querying $f$ for estimated action discrepancy is no longer necessary since we can simply query the robot policy at any state during supervisor mode to obtain a true measure of the action discrepancy between the robot and supervisor policies. This prevents exploiting approximation errors in $f$ when the supervisor is in control. Second, LazyDAgger introduces an asymmetric switching condition between autonomous and supervisor control, where switches are executed from autonomous to supervisor mode if $f$ indicates that the predicted action discrepancy is above $\tausup$, but switches are only executed from supervisor mode back to autonomous mode if the true action discrepancy is below some value $\tauauto < \tausup$. This encourages lengthier interventions, leading to fewer context switches between autonomous and supervisor modes. Finally, LazyDAgger injects noise into supervisor actions in order to spread the distribution of states in which reference controls from the supervisor are available. \algname builds on the asymmetric switching criterion introduced by LazyDAgger, but introduces a new switching criterion based on the estimated task success probability, which we found significantly improved performance in practice.

\begin{algorithm}[t]
\caption{\algname}
\label{alg:main}
\footnotesize
\begin{algorithmic}[1]
\Require Number of episodes $N$, time horizon $T$, supervisor policy $\pisup$, desired context switching rate $\alphasupdesired$
\State Collect offline dataset ${\mathcal{D}_h}$ of $(s, a^h)$ tuples with $\pisup$
\State Initialize $\pirob$ via Behavior Cloning on ${\mathcal{D}_h}$
\State Collect offline dataset ${\mathcal{D}_r}$ of $(s, a^r)$ tuples with $\pirob$
\State Initialize $\safety$ by optimizing Equation~\eqref{eq:RL} on ${\mathcal{D}_r} \cup {\mathcal{D}_h}$
\State Optimize $\tausup, \tauauto, \deltasup, \deltaauto$ on ${\mathcal{D}_h}$ \Comment{Online tuning based on $\alphasupdesired$ (Section~\ref{subsec:tuning})}
\For{$i \in \{1,\ldots N\}$}
    \State Initialize $s_0$, Mode $\gets$ Autonomous
    \For{$t \in \{1,\ldots T\}$}
        \State $a^r_t = \pirob(s_t)$
        \If{Mode = Supervisor or $\textrm{Intervene}(s_t, \deltasup, \tausup)$} \Comment{Determine control mode (Section~\ref{subsec:switches})}
            \State $a^h_t = \pisup(s_t)$
            \State $\mathcal{D}_h \leftarrow \mathcal{D}_h \cup \{(s_t, a_t^h)\}$
            \State Execute $a^h_t$
            \If {$\textrm{Cede}(s_t, \deltaauto, \tauauto)$} \Comment{Default control mode for next timestep (Section 4.3)}
                \State Mode $\gets$ Autonomous
            \Else
                \State Mode $\gets$ Supervisor
            \EndIf
        \Else
            \State Execute $a^r_t$
            \State $\mathcal{D}_r \leftarrow \mathcal{D}_r \cup \{(s_t, a_t)\}$
        \EndIf
        \If{Terminal state reached}
            \State Exit Loop, Mode $\gets$ Autonomous
            \State Recompute $\tausup, \tauauto, \deltasup$ \Comment{Online tuning based on $\alphasupdesired$ (Section~\ref{subsec:tuning})}
        \EndIf
    \EndFor
    \State $\pirob \leftarrow \arg\min_{\pirob} \mathbb{E}_{(s_t, a^h_t)\sim\mathcal{D}_h}\left[\cloningloss\right]$ 
    \State Collect ${\mathcal{D}_r}$ offline with robot policy $\pirob$ 
    \State Update $\safety$ on ${\mathcal{D}_r} \cup {\mathcal{D}_h}$ \Comment{Update Q-function via Equation~\eqref{eq:safetycritic}}
\EndFor
\end{algorithmic}
\end{algorithm}

\subsubsection{HG-DAgger}
Unlike SafeDAgger, LazyDAgger, and \algname, which are robot-gated and autonomously determine when to solicit intervention requests, HG-DAgger is human-gated, and thus requires that the supervisor determine the timing and length of interventions. As in \algname, HG-DAgger updates $\pirob$ on an aggregated dataset of all transitions collected by the supervisor (analogous to $\mathcal{D}_h$ in Algorithm~\ref{alg:main}).

\subsection{Hyperparameter and Implementation Details}
\label{subsec:app_impl_details}
Here we provide a detailed overview of all hyperparameter and implementation details for \algname and all comparisons to facilitate reproduction of all experiments. We also include code in the supplement, and will release a full open-source codebase after anonymous review.

\subsubsection{\algname}
\paragraph{Peg Insertion (Simulation): }We initially populate $\mathcal{D}_h$ with 2,687 offline transitions, which correspond to 30 task demonstrations collected by an expert human supervisor, to initialize the robot policy $\pirob$. We represent $\pirob$ with an ensemble of 5 neural networks, trained on bootstrapped samples of data from $\mathcal{D}_h$ in order to quantify uncertainty for novelty estimation. Each neural network is trained using the Adam Optimizer (learning rate $1\mathrm{e}{-3}$) with 5 training epochs, 500 gradient steps in each training epoch, and a batch size of 100. All networks consist of 2 hidden layers, each with 256 hidden units with ReLU activations, and a Tanh output activation.

The Q-function used for risk-estimation, $\safety$, is trained with a batch size of 50, and batches are balanced such that 10\% of all sampled transitions contain a state in the goal set. We train $\safety$ with the Adam Optimizer, with a learning rate of $1\mathrm{e}{-3}$ and discount factor $\gamma=0.9999$. In order to train $\safety$, we collect 10 test episodes from $\pirob$ every 2,000 environment steps. We represent $\safety$ with a 2 hidden layer neural net in which each hidden layer has 256 hidden units with ReLU activations and with a sigmoid output activation. The state and action are concatenated before they are fed into $\safety$.

\paragraph{Block Stacking (Simulation): } This is an additional simulation environment not included in the main text. Results and a description of the task are in Section~\ref{ssec:block}. We populate $\mathcal{D}_h$ with 1,677 offline transitions, corresponding to 30 task demonstrations, to initialize $\pirob$. All other parameters and implementation details are identical to the peg insertion environment.

\paragraph{Cable Routing (Physical): }We initially populate $\mathcal{D}_h$ with 1,381 offline transitions, corresponding to 25 task demonstrations collected by an expert human supervisor, to initialize the robot policy $\pirob$. We again represent $\pirob$ with an ensemble of 5 neural networks, trained on bootstrapped samples of data from $\mathcal{D}_h$ in order to quantify uncertainty for novelty estimation. Each neural network is trained using the Adam Optimizer (learning rate $2.5\mathrm{e}{-4}$) with 5 training epochs, 300 gradient steps per training epoch, and a batch size of 64. All networks consist of 5 convolutional layers (format: $(\text{in}\_\text{channels}, \text{out}\_\text{channels}, \text{kernel}\_\text{size}, \text{stride})$): $[(3,24,5,2), (24,36,5,2), (36,48,5,2), (48,64,3,1), (64,64,3,1)]$ followed by 4 fully connected layers (format: ($\text{in}\_\text{units}, \text{out}\_\text{units})$): $[(64,100), (100,50), (50,10), (10,2)]$. Here we utilize ELU (exponential linear unit) activations with a Tanh output activation. 

The Q-function used for risk-estimation, $\safety$, is trained with a batch size of 64  as well, and batches are balanced such that 10\% of all sampled transitions contain a state in the goal set.  We train $\safety$ with the Adam Optimizer with a learning rate of $2.5\mathrm{e}{-4}$ and discount factor $\gamma=0.9999$. In order to train $\safety$, we collect 5 test episodes from $\pirob$ every 500 environment steps. We represent $\safety$ with a neural network with the same 5 convolutional layers as the policy networks above, but with the fully connected layers as follows (format: ($\text{in}\_\text{units}, \text{out}\_\text{units})$): $[(64+2,100), (100,50), (50,10), (10,1)]$. We concatenate the action with the state embedding resulting from the 5 convolutional layers (hence the 64 + 2) and feed the resulting concatenated embedding into the 4 fully connected layers above. We utilize ELU (exponential linear unit) activations with a sigmoid output activation. 

\subsubsection{Behavior Cloning}
\paragraph{Peg Insertion (Simulation): }For Behavior Cloning, we initially populate $\mathcal{D}_h$ with 4,004 offline transitions, corresponding to 45 task demonstrations collected by an expert human supervisor, to initialize the robot policy $\pirob$ (note that this is more transitions than are provided to \algname). All other details are the same as \algname for training $\pirob$.

\paragraph{Block Stacking (Simulation): }\new{We initially populate $\mathcal{D}_h$ with 3,532 offline transitions, corresponding to 60 task demonstrations, to initialize $\pirob$. Note that Behavior Cloning has access to twice as many offline demonstrations as the other algorithms.}

\paragraph{Cable Routing (Physical): } We train  $\pirob$ with the same architecture and procedure as for \algname, but only on the initial offline data.

\subsubsection{SafeDAgger}

We use the same hyperparameters and architecture for training $\pirob$ as for \algname. Unlike \algname, SafeDAgger does not have a mechanism to \new{automatically set intervention thresholds} when provided an intervention budget $\alphasupdesired$. Thus, we must specify a value for the switching threshold $\tausup$. We use $\tausup=0.008$, since this is recommended in~\cite{safe_dagger} as the value which was found to work well in experiments (in practice, this value marks about 20\% of states as ``unsafe").

\subsubsection{LazyDAgger}

We use the same hyperparameters and architecture for training $\pirob$ as for \algname. Unlike \algname, LazyDAgger does not have a mechanism to \new{automatically set intervention thresholds} when provided an intervention budget $\alphasupdesired$. Thus, we must specify a value for both switching thresholds $\tausup$ and $\tauauto$. We use $\tausup = 0.015$, $\tauauto = 0.25\tausup$ and use a noise covariance matrix of $0.02\mathcal{N}(0, I)$ when injecting noise into the supervisor actions. These values were tuned to strike a balance between supervisor burden and policy performance.

\subsubsection{HG-DAgger}
All hyperparameters and architectures are identical to those used for Behavior Cloning, without the extra offline demonstrations. Note that for HG-DAgger, the supervisor determines the timing and length of interventions.

\subsection{Environment Details \new{and Additional Metrics}}\label{subsec:app_exp_details}

\subsubsection{Peg Insertion in Simulation}
We evaluate our algorithm and baselines in the Robosuite environment (\url{https://robosuite.ai})~\cite{robosuite}, which builds on MuJoCo~\cite{mujoco} to provide a standardized suite of benchmark tasks for robot learning. Specifically, we consider the ``Nut Assembly" task, in which a robot must grab a ring from a random initial pose and place it over a cylinder at a fixed location. We consider a variant of the task that considers only 1 ring and 1 target, though the simulator allows 2 rings and 2 targets. The states are $s \in \mathbb{R}^{51}$ and actions $a \in \mathbb{R}^5$ (translation in the XY-plane, translation in the Z-axis, rotation around the Z-axis, and opening or closing the gripper). The simulated robot arm is a UR5e, and the controller reaches a commanded pose via operational space control with fixed impedance. To avoid bias due to variable teleoperation speeds \new{and ensure that the Markov property applies}, we abstract 10 timesteps in the simulator into 1 environment step, and in supervisor mode we pause the simulation until keyboard input is received. \new{This prevents accidentally collecting ``no-op" expert labels and allows the end effector to ``settle" instead of letting its momentum carry on to the next state. In practice it does not make the task more difficult, as control is still fine-grained enough for precise manipulation.} Each episode is terminated upon successful task completion or when 175 actions are executed. Interventions are collected through a keyboard interface. \new{In Table~\ref{tab:insertion-extra}, we report additional metrics for the peg insertion simulation experiment and find that \algname solicits fewer interventions than prior algorithms at training time while achieving a higher success rate during training than all algorithms other than HG-DAgger, though it does request more human actions. The train success rate column also indicates that \algname achieves \textit{throughput} comparable to HG-DAgger and higher than other baselines, as \algname has more task successes in the same amount of time (10,000 timesteps for all algorithms). At execution time, \algname collects lengthier interventions than prior algorithms, but as a result is able to succeed more often at execution time as discussed in the main text.}

\begin{table}[t]
\caption{\new{\textbf{Peg Insertion in Simulation Additional Metrics: } We report additional statistics for the peg insertion task: total number of interventions (T Ints), total number of offline and online human actions (T Acts (H)), and total number of robot actions (T Acts (R))) at training time across all trajectories (successful and unsuccessful). We report these same metrics at execution time, but T Acts (H) does not include offline human actions, as at execution time it does not refer to the number of training samples for the robot policy. We also report the success rate of the mixed control policy at \textit{training} time (Train Succ.). Results suggest that \algname solicits fewer interventions than prior algorithms at training time while achieving a comparable success rate and throughput to HG-DAgger. At execution time, \algname collects lengthier interventions than prior algorithms but succeeds more often at the task (Table~\ref{tab:sim}).}}
\centering
\resizebox{\columnwidth}{!}{%
{\begin{tabular}{|l||r|r|r||r||r|r|r||r||}
    \hline
    \multicolumn{1}{|c||}{Algorithm} &
    \multicolumn{3}{|c||}{Training Interventions} &
    \multicolumn{1}{|c||}{Train Succ.} & 
    \multicolumn{3}{|c||}{Execution Interventions}\\
    \cline{2-4}
    \cline{6-8}
      \multicolumn{1}{|c||}{} &
      \multicolumn{1}{|c|}{T Ints} &
      \multicolumn{1}{|c|}{T Acts (H)} &
      \multicolumn{1}{|c||}{T Acts (R)} &
      \multicolumn{1}{|c||}{} &
      \multicolumn{1}{|c|}{T Ints} &
      \multicolumn{1}{|c|}{T Acts (H)} &
      \multicolumn{1}{|c||}{T Acts (R)}\\
    \hline
    \cline{1-8}
     Behavior Cloning & N/A & 4004 & N/A & N/A & N/A & N/A & N/A \\ 
     SafeDAgger & 334 & 4227 & 8460 & $48/73$ & 81 & 396 & 1781 \\ 
     LazyDAgger & 82 & 3683 & 9004 & $37/67$ & 30 & 290 & 2422 \\ 
     HG-DAgger & 124 & 4392 & 8295 & $83/83$ & 23 & 342 & 2071 \\ 
     Ours (-Novelty) & 60 & 5242 & 7445 & $62/80$ & 12 & 157 & 2649 \\ 
     Ours (-Risk)  & 87 & 3623 & 9064 & $72/81$ & 30 & 237 & 2255 \\ 
     Ours: \algname & 84 & 6840 & 5847 & $76/86$ & 27 & 426 & 1696 \\
    \hline
\end{tabular}}}
\label{tab:insertion-extra}
\end{table}
\subsubsection{Block Stacking in Simulation}\label{ssec:block}
\begin{figure}
    \centering
    \includegraphics[width=0.9\textwidth]{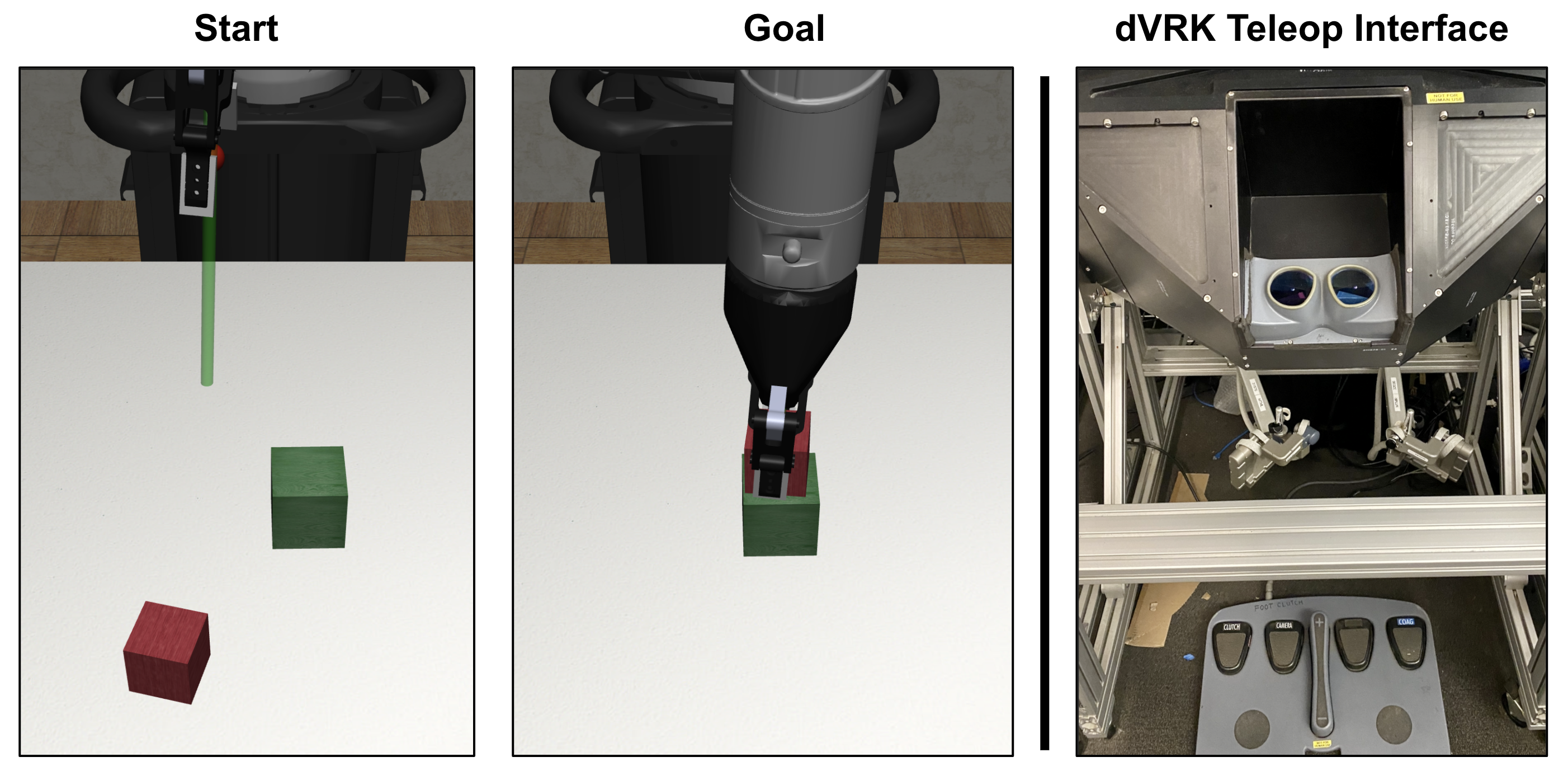}
    \caption{\new{\textbf{Left:} An example start and goal state for the block stacking environment in Robosuite. The goal is to place the red block on top of the green one. Initial poses of both blocks are randomized. \textbf{Right:} The da Vinci Research Kit Master Tool Manipulator (MTM) 7DOF interface used to provide human interventions in the physical experiments. The human expert views the workspace through the viewer (top) and teleoperates the robot by moving the right joystick (middle) in free space while pressing the rightmost pedal (bottom).}}
    \label{fig:stacking}
\end{figure}

\new{To further evaluate the algorithm and baselines in simulation, we also consider the block stacking task from Robosuite (see previous section). Here the robot must grasp a cube in a randomized initial pose and place it on top of a second cube in another randomized pose. See Table~\ref{tab:stack} for training results and Figure~\ref{fig:stacking} for an illustration of the experimental setup. Due to the randomized place position, small placement region, and geometric symmetries, the task is more difficult than peg insertion, as evidenced by the lower autonomous success rate for all algorithms. However, we still see that \new{\algname achieves comparable performance to HG-DAgger in terms of autonomous success rate, success rate during training, and throughput, while outperforming the other baselines and ablations. \algname also solicits fewer interventions than prior algorithms, but generally requires more human actions as these interventions tend to be lengthier. This makes \algname well-suited to situations in which the cost of context switches (latency) may be high.}
}

\begin{table}[t]
\caption{\new{\textbf{Block Stacking in Simulation Results: } We report the number of interventions (Ints), number of human actions (Acts (H)), and number of robot actions (Acts (R)) during training (over successful trajectories as in Table~\ref{tab:sim}) and report the success rate of the robot policy after training when no interventions are allowed (Auto Succ.). We also report the total number of interventions (T Int), total number of actions from the human (offline and online, in T Acts (H)), total number of actions executed by the robot (T Acts (R)), and the success rate of the mixed control policy during training (Train Succ.). Results suggest that \algname achieves comparable performance to HG-DAgger in terms of both autonomous and training success rates while outperforming the other baselines and ablations. \algname also solicits fewer interventions than prior algorithms, but generally requires more human actions.}}
\centering
\resizebox{\columnwidth}{!}{%
{\begin{tabular}{|l|r|r|r|r|r|r|r|r|}
    \hline
     Algorithm & Ints & Acts (H) & Acts (R) & Auto Succ. & T Ints & T Acts (H) & T Acts (R) & Train Succ. \\
     \hline
     Behavior Cloning & N/A & N/A & $68.0 \pm 11.4$ & $5/100$ & N/A & 3532 & N/A & N/A \\ 
     SafeDAgger & $5.00 \pm 3.41$ & $40.5 \pm 14.1$ & $44.3 \pm 25.6$ & $3/100$ & 574 & 4387 & 7290 & 27/68\\ 
     LazyDAgger & $1.81 \pm 1.02$ & $25.8 \pm 17.8$ & $56.6 \pm 28.3$ & $40/100$ & 85 & 2940 & 8737 & 36/75 \\ 
     HG-DAgger & $1.62 \pm 0.91$ & $22.5 \pm 16.5$ & $54.6 \pm 14.2$ & $\mathbf{56/100}$ & 201 & 4535 & 7142 & 124/125 \\ 
     Ours (-Novelty) & $0.65 \pm 0.70$ & $43.7 \pm 13.3$ & $28.6 \pm 28.5$ & $8/100$ & 37 & 3599 & 8078 & 23/69 \\ 
     Ours (-Risk)  & $1.89 \pm 0.72$ & $12.9 \pm 7.7$ & $72.4 \pm 25.5$ & $31/100$ & 109 & 2518 & 9159 & 47/79 \\
     Ours: \algname & $1.33 \pm 0.76$ & $35.4 \pm 15.8$ & $37.2 \pm 27.5$ & $\mathbf{52/100}$ & 153 & 5873 & 5804 & 111/120 \\
    \hline
\end{tabular}}}
\label{tab:stack}
\end{table}

\subsubsection{Physical Cable Routing}
Finally, we evaluate our algorithm on a visuomotor cable routing task with a da Vinci Research Kit surgical robot. We take RGB images of the scene with a Zivid One Plus camera inclined at about 45 degrees to the vertical. These images are cropped into a square and downsampled to 64 $\times$ 64 before they are passed to the neural network policy. The cable state is initialized to approximately the same shape (see Figure~\ref{fig:vis}) with the cable initialized in the robot's gripper. The workspace is approximately 10 cm $\times$ 10 cm, and each component of the robot action ($\Delta x, \Delta y$) is at most 1 cm in magnitude. To avoid collision with the 4 obstacles, we implement a ``logical obstacle" as 1-cm radius balls around the center of each obstacle. Actions that enter one of these regions are projected to the boundary of the circle. Each episode is terminated upon successful task completion or 100 actions executed. \new{Interventions are collected through a 7DOF teleoperation interface (Figure~\ref{fig:stacking}) that matches the pose of the robot arm, with rotation of the end effector disabled. Teleoperated actions are mapped to the robot's action space by projecting pose deltas to the 2D plane at 1 second intervals. The human teleoperates the robot at a frequency that roughly corresponds to taking actions with the maximum magnitude (1 cm / sec).} \new{In Table~\ref{tab:cable-extra}, we report additional metrics for the physical cable routing experiment and find that \algname solicits a number of interventions similar to HG-DAgger while achieving a similar success rate during training. This again indicates that \algname is able to learn intervention criteria competitive with human judgment. At execution time, we find that \algname solicits the same number of interventions as HG-DAgger, but requires fewer human and robot actions than HG-DAgger.}

\begin{table}[t]
\caption{\new{\textbf{Physical Cable Routing Additional Metrics: } We report additional statistics for the peg insertion task: total number of interventions (T Ints), total number of offline and online human actions (T Acts (H)), and total number of robot actions (T Acts (R))) at training time across all trajectories. We report these same metrics at execution time, but T Acts (H) does not include offline human actions, as at execution time it does not refer to the number of training samples for the robot policy. We also report the success rate of the mixed control policy at \textit{training} time (Train Succ.). Results suggest that \algname needs fewer interventions than HG-DAgger while achieving a similar training success rate. At execution time, we find that \algname solicits the same number of interventions as HG-DAgger, but requires fewer human and robot actions.}}
\centering
\resizebox{\columnwidth}{!}{%
{\begin{tabular}{|l||r|r|r||r||r|r|r||r||}
    \hline
    \multicolumn{1}{|c||}{Algorithm} &
    \multicolumn{3}{|c||}{Training Interventions} &
    \multicolumn{1}{|c||}{Train Succ.} & 
    \multicolumn{3}{|c||}{Execution Interventions}\\
    \cline{2-4}
    \cline{6-8}
      \multicolumn{1}{|c||}{} &
      \multicolumn{1}{|c|}{T Ints} &
      \multicolumn{1}{|c|}{T Acts (H)} &
      \multicolumn{1}{|c||}{T Acts (R)} &
      \multicolumn{1}{|c||}{} &
      \multicolumn{1}{|c|}{T Ints} &
      \multicolumn{1}{|c|}{T Acts (H)} &
      \multicolumn{1}{|c||}{T Acts (R)}\\
    \hline
    \cline{1-8}
     Behavior Cloning & N/A & 1381 & N/A & N/A & N/A & N/A & N/A \\ 
     HG-DAgger & 31 & 1682 & 1199 & $20/20$ & 6 & 41 & 1109 \\ 
     Ours: \algname & 27 & 1728 & 1153 & $19/21$ & 6 & 23 & 919 \\
    \hline
\end{tabular}}}
\label{tab:cable-extra}
\end{table}

\subsection{User Study Details}
\label{subsec:app_user_study}
Here we detail the protocol for conducting user studies with \algname and comparisons and discuss qualitative results based on participants' answers to survey questions measuring their mental load and levels of frustration when using each of the algorithms.

\subsubsection{User Study Interface}
\begin{figure*}
    \centering
    \includegraphics[width=\textwidth]{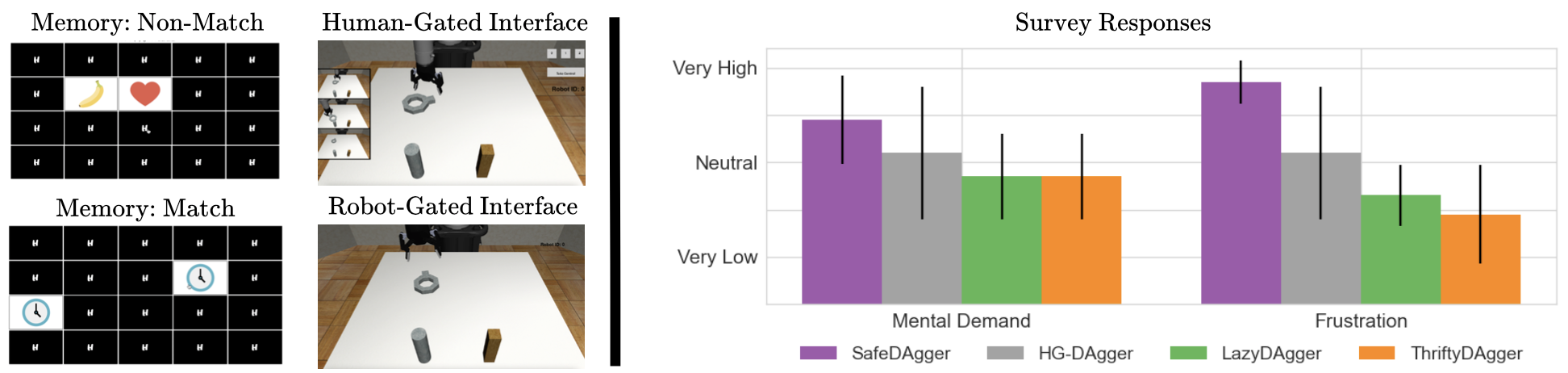}
    \caption{\textbf{User Study Survey Results: }We illustrate the user study interface for the human-gated and robot-gated algorithms (left) and users' survey responses regarding their mental load and frustration (right) for each algorithm. Results suggest that users report similar levels of mental load and frustration for \algname and LazyDAgger, but significantly higher levels of both metrics for HG-DAgger and SafeDAgger. We hypothesize that the sparing and sustained interventions solicited by \algname and LazyDAgger lead to greater user satisfaction and comfort compared to algorithms which force the user to constantly monitor the system (HG-DAgger) or frequently context switch between teleoperation and the distractor task.}
    \label{fig:survey_results}
\end{figure*}

Figure~\ref{fig:survey_results} (left) illustrates the interface used for the user study. The user study is performed with the same peg insertion environment used for simulation experiments, but with 3 robots performing the task in parallel. The base policy is initialized from 30 demos, as in the other simulation experiments. To speed up the task execution for the user study, each action has twice the magnitude as in the peg insertion experiments. Since this results in shorter trajectories that are easier for Behavior Cloning to accomplish, we also inject a small amount of Gaussian noise (covariance matrix = $0.02\mathcal{N}(0, I)$).

In the human-gated study with HG-DAgger, participants are shown videos of all 3 robots attempting to perform the task in a side pane (Figure~\ref{fig:survey_results}, top right of left pane) and are instructed to monitor all of the robots and intervene when they deem it appropriate. In all robot-gated studies, participants are instructed to play the Concentration game until they hear a chime, at which point they are instructed to switch screens to the teleoperation interface. The Concentration game (also called Memory) is illustrated on the left of the left pane in Figure~\ref{fig:survey_results}: the objective is to find pairs of cards (all of which are initially face-down) which have matching pictures on their front side. Examples of a non-matching pair and a matching pair are illustrated in Figure~\ref{fig:survey_results}.

All robots which require interventions are placed in a FIFO queue, with participants serving intervention requests sequentially until no robot requires intervention. Thus, the participant may be required to provide interventions for multiple robots in succession if multiple robots are currently in the queue. When no robot requires assistance, the teleoperation interface turns black and reports that no robot currently needs help, at which point participants are instructed to return to the Concentration game. 

\subsubsection{NASA TLX Survey Results}
After each participant is subjected to all 4 conditions (SafeDAgger, LazyDAgger, ThriftyDAgger, and HG-DAgger) in a randomized order, we give each participant a NASA TLX survey asking them to rate their mental demand and frustration for each of the conditions on a scale of 1 (very low) - 5 (very high). Results (Figure~\ref{fig:survey_results} right pane) suggest that ThriftyDAgger and LazyDAgger impose less mental demand and make participants feel less frustrated than HG-DAgger and SafeDAgger. During experiments, we found that participants found it cumbersome to keep track of all of the robots simultaneously in HG-DAgger, while the frequent context switches in SafeDAgger made participants frustrated since they were often unable to make much progress in the Concentration Game and felt that the robot repeatedly asked for interventions in very similar states.

\subsubsection{Wall Clock Time}
\new{We report additional metrics on the wall clock time of each condition in Table~\ref{tab:wallclocktime}. Since all experiments are run for the same 350 time steps, total wall clock time is relatively consistent. However, HG-DAgger takes longer, as it takes more compute to render all three robot views at once. \algname takes less total human time than the baselines, allowing the human to make more progress on independent tasks. Note that other robots in autonomous mode can still make task progress during human intervention. Note also that HG-DAgger requires human attention for the Total Wall Clock Time, as the human must supervise all the robots even if he or she is not actively teleoperating one (as recorded by Human Wall Clock Time).}

\begin{table}[t]
\caption{\new{\textbf{Wall Clock Time: } We compare the total amount of wall clock time and total amount of human wall clock time averaged over the 10 subjects in the user study. Human Wall Clock Time refers to the amount of time the human spent actively teleoperating a robot, while Total Wall Clock Time measures the amount of time taken by the total experiment. \algname requires the lowest amount of human time, and the total amount of time is relatively consistent. Note that HG-DAgger takes more Total Wall Clock Time as it takes longer to simulate the ``bird's eye view" of all 3 robots, and that autonomous robots can still make task progress independently while a human is operating a robot.}}
\centering
\resizebox{0.75\columnwidth}{!}{%
{\begin{tabular}{|l|r|r|}
    \hline
     Algorithm & Human Wall Clock Time (s) & Total Wall Clock Time (s) \\
     \hline
     SafeDAgger & $448.0 \pm 48.1$ & $613.0 \pm 33.1$  \\ 
     LazyDAgger & $415.3 \pm 90.3$ & $609.6 \pm 49.5$ \\ 
     HG-DAgger & $532.6 \pm 105.2$ & $792.8 \pm 68.7$ \\ 
     Ours: \algname & $\mathbf{365.4 \pm 88.1}$ & $625.5 \pm 52.3$ \\
    \hline
\end{tabular}}}
\label{tab:wallclocktime}
\end{table}

\subsubsection{Detailed Protocol}
For the user study, we recruited 10 participants aged 18-37, including members without any knowledge or experience in robotics or AI. All participants are first assigned a randomly selected user ID. Then, participants are instructed to play a 12-card game of Concentration (also known as Memory) (\url{https://www.helpfulgames.com/subjects/brain-training/memory.html}) in order to learn how to play. Then, users are given practice with both the robot-gated and human-gated teleoperation interfaces. To do this, the operator of the study (one of the authors) performs one episode of the task in the robot-gated interface and briefly explains how to control the human-gated interface. Then, participants are instructed to perform one practice episode in the robot-gated teleoperation interface and spend a few minutes exploring the human-gated interface until they are confident in the usage of both interfaces and in how to teleoperate the robots. In the robot-gated experiments, participants are instructed to play Concentration when no robot asks for help, but to immediately switch to helping the robot whenever a robot asks for help. In the human-gated experiment with HG-DAgger, participants are instructed to continuously monitor all of the robots and perform interventions which they believe will maximize the number of successful episodes. During the robot-gated study, participants play the 24-card version of Concentration between robot interventions. If a participant completes the game, new games of Concentration are created until a time budget of robot interactions is hit. Then for each condition, the participant is scored based on (1) the number of times the robot successfully completed the task and (2) the number of total matching pairs the participant found across all games of Concentration.

\end{document}